# Generalized Earley Parser: Bridging Symbolic Grammars and Sequence Data for Future Prediction

**Siyuan Qi** [1] **Baoxiong Jia** [1 2] **Song-Chun Zhu** [1]

## Abstract

Future predictions on sequence data (*e.g.*, videos or audios) require the algorithms to capture non-Markovian and compositional properties of high-level semantics. Context-free grammars are natural choices to capture such properties, but traditional grammar parsers (*e.g.*, Earley parser) only take symbolic sentences as inputs. In this paper, we generalize the Earley parser to parse sequence data which is neither segmented nor labeled. This generalized Earley parser integrates a grammar parser with a classifier to find the optimal segmentation and labels, and makes *top-down* future predictions. Experiments show that our method significantly outperforms other approaches for future human activity prediction.

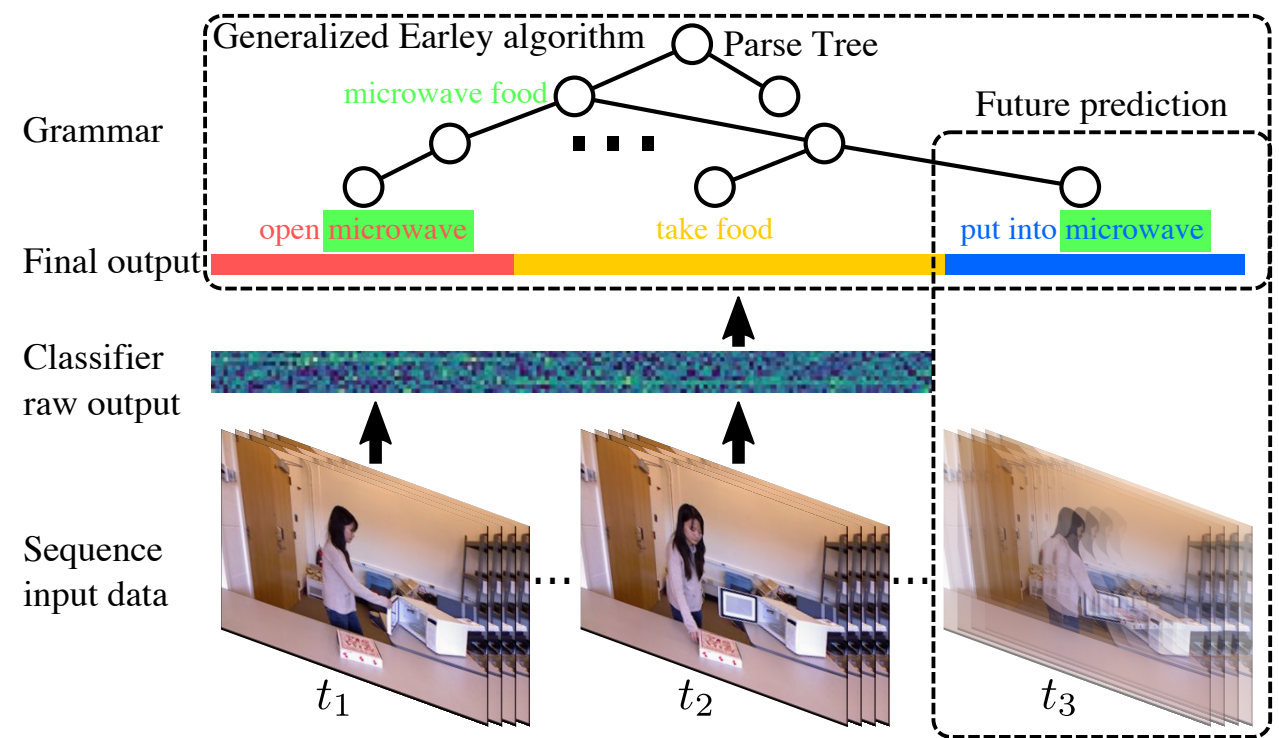

*Figure 1.* The input of the generalized Earley parser is a matrix of probabilities of each label for each frame, given by an arbitrary classifier. The parser segments and labels the sequence data into a label sentence in the language of a given grammar. Future predictions are then made based on the grammar.

## 1. Introduction

We consider the task of labeling unsegmented sequence data and predicting future labels. This is a ubiquitous problem with important applications in many perceptual tasks, *e.g.*, recognition and anticipation of human activities or speech. A generic modeling approach is key for intelligent agents to perform future-aware tasks (*e.g.*, assistive activities).

Such tasks often exhibit non-Markovian and compositional properties, which should be captured by a top-down prediction algorithm. Consider the video sequence in Figure 1, human observers recognize the first two actions and predict the most likely future action based on the entire history. Context-free grammars are natural choices to model such reasoning processes, and it is one step closer to Turing machines than Markov models (*e.g.*, Hidden Markov Models) in the Chomsky hierarchy.

However, it has not been possible to directly use symbolic grammars to parse and label sequence data. Traditional grammar parsers take symbolic sentences as inputs instead of noisy sequence data like videos or audios. The data has to be i) segmented and ii) labeled to apply existing grammar parsers to. One naive solution is to first segment and label the data using a classifier and thus generating a label sentence. Then grammar parsers can be applied on top of it for prediction. But this is apparently non-optimal, since the grammar rules are not considered in the classification process. It may not even be possible to parse this label sentence, because they are very often grammatically incorrect.

In this paper, we design a grammar-based parsing algorithm that directly operates on sequence input data, which goes beyond the scope of symbolic string inputs. Specifically, we propose a generalized Earley parser based on the Earley parser (Earley, 1970). The algorithm finds the optimal segmentation and label sentence according to both a symbolic grammar and a classifier output of probabilities of labels for each frame as shown in Figure 1. Optimality here means maximizing the probability of the label sentence according to the classifier output while being *grammatically correct*.

The difficulty of achieving this optimality lies in the joint optimization of both the grammar structure and the parsing likelihood of the output label sentence. For example, an expectation-maximization-type algorithm will not work well

[1]University of California, Los Angeles, USA [2]Peking University, Beijing, China. Correspondence to: Siyuan Qi <syqi@cs.ucla.edu>.

*Proceedings of the $35^{th}$ International Conference on Machine Learning*, Stockholm, Sweden, PMLR 80, 2018.

since i) there is no guarantee for optimality, and ii) any grammatically incorrect sentence has a grammar prior of probability 0. The algorithm can easily get stuck in local minimums and fail to find a grammatically correct solution.

The core idea of our algorithm is to directly and efficiently search for the optimal label sentence in the language defined by the grammar. The constraint of the search space ensures that the sentence is grammatically correct. Specifically, a heuristic search is performed on the prefix tree expanded according to the grammar, where the path from the root to a node represents a partial sentence (prefix). By carefully defining the heuristic as a prefix probability computed based on the classifier output, we can efficiently search through the tree to find the optimal label sentence.

The generalized Earley parser has four major **advantages**. **i)** The inference process highly integrates a high-level grammar with an underlying classifier; the grammar gives guidance for segmenting and labeling the sequence data. **ii)** It can be applied to any classifier outputs. **iii)** It generates a grammar parse tree for data sequence that is highly explainable. **iv)** It is principled and generic, as it applies to most sequence data parsing and prediction problems.

We evaluate the proposed approach on two datasets of human activities in the computer vision domain. The first dataset CAD-120 (Koppula et al., 2013) consists of daily activities and most activity prediction methods are based on this dataset. Comparisons show that our method significantly outperforms state-of-the-art methods on future activity prediction. The second dataset Watch-n-Patch (Wu et al., 2015) is designed for "action patching", which includes daily activities that have action forgotten by people. Experiments on the second dataset show the robustness of our method on noisy data. Both experiments show that the generalized Earley parser performs particularly well on prediction tasks, primarily due to its non-Markovian property.

This paper makes three major **contributions**.
• We design a parsing algorithm for symbolic context-free grammars. It directly operates on sequence data to obtain the optimal segmentation and labels.
• We propose a prediction algorithm that naturally integrates with this parsing algorithm.
• We formulate an objective for future prediction for both grammar induction and classifier training. The generalized Earley parser serves as a concrete example for combining symbolic reasoning methods and connectionist approaches.

## 2. Background: Context-Free Grammars

In formal language theory, a context-free grammar is a type of formal grammar, which contains a set of production rules that describe all possible sentences in a given formal language. A context-free grammar $G$ in Chomsky Normal Form is defined by a 4-tuple $G = (V, \Sigma, R, \gamma)$ where
1. $V$ is a finite set of non-terminal symbols that can be replaced by/expanded to a sequence of symbols.
2. $\Sigma$ is a finite set of terminal symbols that represent actual words in a language, which cannot be further expanded.
3. $R$ is a finite set of production rules describing the replacement of symbols, typically of the form $A \rightarrow BC$ or $A \rightarrow \alpha$ for $A, B, C \in V$ and $\alpha \in \Sigma$.
4. $\gamma \in V$ is the start symbol (root of the grammar).

Given a formal grammar, parsing is the process of analyzing a string of symbols, conforming to the production rules and generating a parse tree. A parse tree represents the syntactic structure of a string according to some context-free grammar. The root node of the tree is the grammar root. Other non-leaf nodes correspond to non-terminals in the grammar, expanded according to grammar production rules. The leaf nodes are terminal symbols. All the leaf nodes together form a sentence.

The above definition can be augmented by assigning a probability to each production rule, thus becoming a probabilistic context-free grammar. The probability of a parse tree is the product of the production rules that derive the parse tree.

Sample grammar: $\gamma \rightarrow R$; $R \rightarrow R + R$; $R \rightarrow$ "0" | "1"
Input string: $0 + 1$
State: | state # | rule | origin | comment |

| | | | |
|---|---|---|---|
| $S(0)$ | | | |
| (1) | $\gamma \rightarrow \cdot R$ | 0 | start rule |
| (2) | $R \rightarrow \cdot R + R$ | 0 | predict: (1) |
| (3) | $R \rightarrow \cdot 0$ | 0 | predict: (1) |
| (4) | $R \rightarrow \cdot 1$ | 0 | predict: (1) |
| $S(1)$ | | | |
| (1) | $R \rightarrow 0 \cdot$ | 0 | scan: $S(0)(3)$ |
| (2) | $R \rightarrow R \cdot + R$ | 0 | complete: (1) and $S(0)(2)$ |
| (3) | $\gamma \rightarrow R \cdot$ | 0 | complete: (2) and $S(0)(1)$ |
| $S(2)$ | | | |
| (1) | $R \rightarrow R + \cdot R$ | 0 | scan: $S(1)(2)$ |
| (2) | $R \rightarrow \cdot R + R$ | 2 | predict: (1) |
| (3) | $R \rightarrow \cdot 0$ | 2 | predict: (1) |
| (4) | $R \rightarrow \cdot 1$ | 2 | predict: (1) |
| $S(3)$ | | | |
| (1) | $R \rightarrow 1 \cdot$ | 2 | scan: $S(2)(4)$ |
| (2) | $R \rightarrow R + R \cdot$ | 0 | complete: (1) and $S(2)(1)$ |
| (3) | $R \rightarrow R \cdot + R$ | 0 | complete: (1) and $S(2)(2)$ |
| (4) | $\gamma \rightarrow R \cdot$ | 0 | complete: (2) and $S(0)(1)$ |

*Figure 2.* An example of the original Earley parser.

## 3. Earley Parser

In this section, we review the original Earley parser (Earley, 1970) and introduce the basic concepts. Earley parser is an algorithm for parsing sentences of a given context-free language. In the following descriptions, $\alpha$, $\beta$, and $\gamma$ represent any string of terminals/nonterminals (including the empty string $\epsilon$), $A$ and $B$ represent single nonterminals, and $a$ represents a terminal symbol. We adopt Earley's dot notation: for production of form $A \rightarrow \alpha\beta$, the notation $A \rightarrow \alpha \cdot \beta$ means $\alpha$ has been parsed and $\beta$ is expected.

Input position $n$ is defined as the position after accepting the $n$th token, and input position 0 is the position prior to input. At each input position $m$, the parser generates a state set $S(m)$. Each state is a tuple $(A \to \alpha \cdot \beta, i)$, consisting of

- The production currently being matched $(A \to \alpha\beta)$.
- The dot: the current position in that production.
- The position $i$ in the input at which the matching of this production began: the position of origin.

Seeded with $S(0)$ containing only the top-level rule, the parser then repeatedly executes three operations: prediction, scanning and completion:

- Prediction: for every state in $S(m)$ of the form $(A \to \alpha \cdot B\beta, i)$, where $i$ is the origin position as above, add $(B \to \cdot\gamma, m)$ to $S(m)$ for every production in the grammar with $B$ on the left-hand side (*i.e.*, $B \to \gamma$).
- Scanning: if $a$ is the next symbol in the input stream, for every state in $S(m)$ of the form $(A \to \alpha \cdot a\beta, i)$, add $(A \to \alpha a \cdot \beta, i)$ to $S(m+1)$.
- Completion: for every state in $S(m)$ of the form $(A \to \gamma\cdot, j)$, find states in $S(j)$ of the form $(B \to \alpha \cdot A\beta, i)$ and add $(B \to \alpha A \cdot \beta, i)$ to $S(m)$.

In this process, duplicate states are not added to the state set. These three operations are repeated until no new states can be added to the set. The Earley parser executes in $O(n^2)$ for unambiguous grammars regarding the string length $n$, and $O(n)$ for almost all $LR(k)$ grammars. A simple example is demonstrated in Figure 2.

## 4. Generalized Earley Parser

In this section, we introduce the proposed algorithm. Instead of taking symbolic sentences as input, we aim to design an algorithm that can parse raw sequence data $\boldsymbol{x}$ of length $T$ (*e.g.*, videos or audios) into a sentence $l$ of labels (*e.g.*, actions or words) of length $|l| \leq T$, where each label $k \in \{0, 1, \cdots, K\}$ corresponds to a segment of a sequence. To achieve that, a classifier (*e.g.*, a neural network) is first applied to each sequence $\boldsymbol{x}$ to get a $T \times K$ probability matrix $\boldsymbol{y}$ (*e.g.*, softmax activations of the neural network), with $y_t^k$ representing the probability of frame $t$ being labeled as $k$. The proposed generalized Earley parser takes $\boldsymbol{y}$ as input and outputs the sentence $l^*$ that best explains the data according to a grammar $G$ of Chomsky normal form.

The core idea is to use the original Earley parser to help construct a prefix tree according to the grammar as illustrated in Figure 3. A prefix tree is composed of terminal symbols and terminations that represent ends of sentences. The root node of the tree is the "empty" symbol. The path from the root to any node in the tree represents a partial sentence (prefix). For each prefix, we can compute the probability that the best label sentence starts with this prefix. This probability

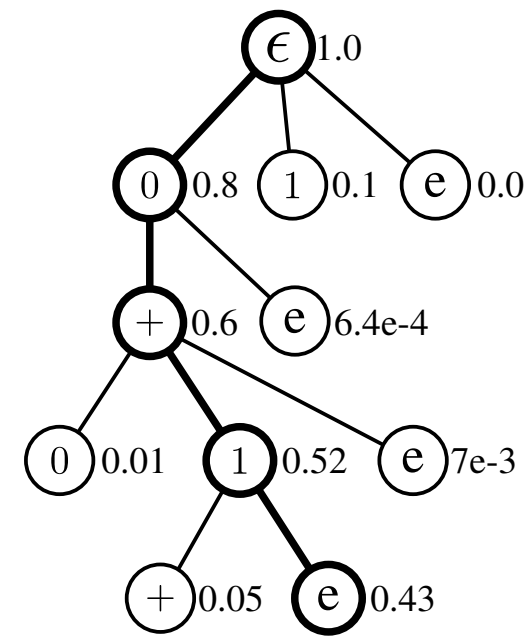

Sample grammar:

$\gamma \to R$

$R \to R + R$

$R \to$ "0" | "1"

Input (classifier output):

| frame | "0" | "1" | "+" |
|---|---|---|---|
| 0 | 0.8 | 0.1 | 0.1 |
| 1 | 0.8 | 0.1 | 0.1 |
| 2 | 0.1 | 0.1 | 0.8 |
| 3 | 0.1 | 0.8 | 0.1 |
| 4 | 0.1 | 0.8 | 0.1 |

*Figure 3.* Prefix search according to grammar. A classifier is applied to a 5-frame signal and outputs a probability matrix (bottom right) as the input to our algorithm. The proposed algorithm expands a grammar prefix tree (left), where "e" represents termination. It finally outputs the best label "$0 + 1$" with probability 0.43. The probabilities of children nodes do not sum to 1 since the grammatically incorrect nodes are eliminated from the search.

is used as a heuristic to search for the best label sentence in the prefix tree: the prefix probabilities prioritize the nodes to be expanded in the prefix tree. The parser finds the best solution when it expands a termination node in the tree. It then returns the current prefix string as the best solution.

This heuristic search is realized by generalizing the Earley parser. Specifically, the scan operation in the Earley parser essentially expands a new node in the grammar prefix tree. For each prefix $l$, we can compute $p(l|x_{0:T})$ and $p(l...|x_{0:T})$ based on $\boldsymbol{y}$, where $p(l|x_{0:T})$ is the probability of $l$ being the best label, and $p(l...|x_{0:T})$ is the probability of $l$ being the prefix of the best label of $x_{0:T}$. The formulations for $p(l|x_{0:T})$ and $p(l...|x_{0:T})$ are derived in Section 4.1.

We now describe the details. Each scan operation will create a new set $S(m,n) \in S(m)$, where $m$ is the length of the scanned string, $n$ is the total number of the terminals that have been scanned at position $m$. This can be thought of as creating a new leaf node in the prefix tree, and $S(m)$ is the set of all created nodes at level $m$. A priority queue $q$ is kept for state sets for prefix search. Scan operations will push the newly created set into the queue with priority $p(l...)$, where $l$ is the parsed string of the state being scanned.

Each state is a tuple $(A \to \alpha \cdot \beta, i, j, l, p(l...))$ augmented from the original Earley parser by adding $j, l, p(l...)$. Here $l$ is the parsed string of the state, and $i, j$ are the indices of the set that this rule originated. The parser then repeatedly executes three operations: prediction, scanning, and completion modified from Earley parser:

- Prediction: for every state in $S(m,n)$ of the form $(A \to \alpha \cdot B\beta, i, j, l, p(l...))$, add $(B \to \cdot\gamma, m, n, l, p(l...))$ to $S(m,n)$ for every production in the grammar with $B$ on the left-hand side.
- Scanning: for every state in $S(m,n)$ of the form $(A \to \alpha \cdot a\beta, i, j, l, p(l...))$, append the new terminal $a$ to $l$ and compute the probability $p((l+a)...)$. Create a new set $S(m+1, n')$ where $n'$ is the current size of $S(m+1)$. Add $(A \rightarrow \alpha a \cdot \beta, i, j, l+a, p((l+a)...))$ to $S(m+1, n')$. Push $S(m+1, n')$ into $q$ with priority $p((l+a)...)$.

- Completion: for every state in $S(m,n)$ of the form $(A \rightarrow \gamma \cdot, i, j, l, p(l...))$, find states in $S(i,j)$ of the form $(B \rightarrow \alpha \cdot A\beta, i', j', l', p(l'...))$ and add $(B \rightarrow \alpha A \cdot \beta, i', j', l, p(l...))$ to $S(m,n)$.

This parsing process is efficient since we do not need to search through the entire tree. As shown in Figure 3 and Algorithm 1, the best label sentence $l$ is returned when the probability of termination is larger than any other prefix probabilities. As long as the prefix probability is computed correctly, it is guaranteed to return the best solution.

The original Earley parser is a special case of the generalized Earley parser. Intuitively, for any input sentence to Earley parser, we can always convert it to one-hot vectors and apply the proposed algorithm. On the other hand, the original Earley parser cannot be applied to segmented one-hot vectors since the labels are often grammatically incorrect. Hence we have the following proposition.

**Proposition 1.** *Earley parser is a special case of the generalized Earley parser.*

*Proof.* Let $L(G)$ denote the language of grammar $G$, $h(\cdot)$ denote a one-to-one mapping from a label to a one-hot vector. $L(G)$ is the input space for Earley parser. $\forall\ l \in L(G)$, the generalized Earley parser accepts $h(l)$ as input. Therefore the proposition follows. □

Here we emphasize two important distinctions of our method to traditional probabilistic parsers with prefix probabilities. i) In traditional parsers, the prefix probability is the probability of a string being a prefix of some strings generated by a grammar (top-down grammar prior). In our case, the parser computes the bottom-up data likelihood. It is straightforward to extend this to a posterior. ii) Traditional parsers only maintain a parse tree, while our algorithm maintains both a parse tree and a prefix tree. The introduction of the prefix tree into the parser enables us to efficiently search in the grammar according to a desired heuristic.

### 4.1. Parsing Probability Formulation

The parsing probability $p(l|x_{0:T})$ is computed in a dynamic programming fashion. Let $k$ be the last label in $l$. For $t = 0$, the probability is initialized by:

$$p(l|x_0) = \begin{cases} y_0^k & l \text{ contains only } k \\ 0 & \text{otherwise} \end{cases} \tag{1}$$

Let $l^-$ be the label sentence obtained by removing the last label $k$ from the label sentence $l$. For $t > 0$, the last frame $t$ must be classified as $k$. The previous frames can be labeled as either $l$ or $l^-$. Then we have:

$$p(l|x_{0:t}) = y_t^k(p(l|x_{0:t-1}) + p(l^-|x_{0:t-1})) \tag{2}$$

**Algorithm 1** Generalized Earley Parser

> Initialize: $S(0,0) = \{(\gamma \rightarrow \cdot R, 0, 0, \epsilon, 1.0)\}$.
> $q = priorityQueue()$
> $q.push(1.0, (0, 0, \epsilon, S(0,0)))$ {Push to queue with prob 1.0 and index (0, 0)}
> **while** $(m, n, l^-, currentSet) = q.pop()$ **do**
>   **for** $s = (r, i, j, l, p(l...)) \in currentSet$ **do**
>     **if** $p(l) > p(l^*)$: $l^* = l$
>     **if** $r$ **is** $(A \rightarrow \alpha \cdot B\beta)$ **then** {***prediction***}
>       **for** each $(B \rightarrow \gamma)$ **in** grammar $g$ **do**
>         $r' = (B \rightarrow \cdot\gamma)$
>         $s' = (r', m, n, l, p(l...))$
>         $S(m,n).add(s')$
>       **end for**
>     **else if** $r$ **is** $(A \rightarrow \alpha \cdot a\beta)$ **then** {***scanning***}
>       $r' = (A \rightarrow \alpha a \cdot \beta)$
>       $m' = m+1, n' = |S(m+1)| - 1$
>       $s' = (r', i, j, l+a, p((l+a)...))$
>       $S(m', n').add(s')$
>       $q.push(p((l+a)...), (m', n', S(m', n')))$
>     **else if** $r$ **is** $(B \rightarrow \gamma \cdot)$ **then** {***completion***}
>       **for** each $((A \rightarrow \alpha \cdot B\beta), i', j')$ **in** $S(i,j)$ **do**
>         $r' = (A \rightarrow \alpha B \cdot \beta)$
>         $s' = (r', i', j', l, p(l...))$
>         $S(m,n).add(s')$
>       **end for**
>     **end if**
>   **end for**
>   **if** $p(l^-) > p(l)$ **for** all un-expanded $l$: **return** $l^-$
> **end while**
> **return** $l^*$

It is worth mentioning that when $y_t^k$ is wrongly given as 0, the dynamic programming process will have trouble correcting the mistake. Even if $p(l^-|x_{0:t-1})$ is high, the probability $p(l|x_{0:t})$ will be 0. Fortunately, since the softmax function is usually adopted to compute $y$, $y_t^k$ will not be 0 and the solution will be kept for consideration.

Then we compute the prefix probability $p(l...|x_{0:T})$ based on $p(l^-|x_{0:t})$. For $l$ to be the prefix, the transition from $l^-$ to $l$ can happen at any frame $t \in \{0, \cdots, T\}$. Once the label $k$ is observed (the transition happens), $l$ becomes the prefix and the rest frames can be labeled arbitrarily. Hence the probability of $l$ being the prefix is:

$$p(l...|x_{0:T}) = p(l|x_0) + \sum_{t=1}^{T} y_t^k p(l^-|x_{0:t-1}) \tag{3}$$

In practice, the probability $p(l|x_{0:t})$ decreases exponentially as $t$ increases and will soon lead to numeric underflow. To avoid this, the probabilities need to be computed in log space. The time complexity of computing the probabilities is $O(T)$ for each sentence $l$ because $p(l^-|x_{0:t})$ are cached. The worst case complexity of the entire parsing is $O(T|G|)$.

### 4.2. Segmentation and Labeling

The generalized Earley parser gives us the best grammatically correct label sentence $l$ to explain the sequence data, which takes all possible segmentations into consideration. Therefore the probability $p(l|x_{0:T})$ is the summation of probabilities of all possible segmentations. Let $p(l|y_{0:e})$ be the probability of the best segmentation based on the classifier output $y$ for sentence $l$. We perform a maximization over different segmentations by dynamic programming to find the best segmentation:

$$p(l|y_{0:e}) = \max_{b<e} p(l^-|y_{0:b}) \prod_{t=b}^{e} y_t^k \tag{4}$$

where $e$ is the time frame that $l$ ends and $b$ is the time frame that $l^-$ ends. The best segmentation can be obtained by backtracing the above probability. Similar to the previous probabilities, this probability needs to be computed in log space as well. The time complexity of the segmentation and labeling is $O(T^2)$.

### 4.3. Future Label Prediction

Given the parsing result $l$, we make grammar-based predictions for the next label $z$ to be observed. The predictions are naturally obtained by the predict operation in the generalized Earley parser.

To predict the next possible symbols at current position $(m, n)$, we search through the states $S(m, n)$ of the form $(X \rightarrow \alpha \cdot z\beta, i, j, l, p(l...))$, where the first symbol $z$ after the current position is a terminal node. The predictions $\Sigma$ are then given by the set of all possible $z$:

$$\Sigma = \{z : \exists s \in S(m, n), s = (X \rightarrow \alpha \cdot z\beta, i, j, l, p(l...))\} \tag{5}$$

The probability of each prediction is then given by the parsing likelihood of the sentence constructed by appending the predicted label $z$ to the current sentence $l$. Assuming that the best prediction corresponds to the best parsing result, the goal is to find the best prediction $z^*$ that maximizes the following conditional probability as parsing likelihood:

$$z^* = \operatorname*{argmax}_{z\in\Sigma} p(z, l|G) \tag{6}$$

For a grammatically complete sentence $u$, the parsing likelihood is simply the Viterbi likelihood (Viterbi, 1967) given by the probabilistic context-free grammar. For an incomplete sentence $l$ of length $|l|$, the parsing likelihood is given by the sum of all the grammatically possible sentences:

$$p(l|G) = \sum_{u_{1:|l|}=l} p(u|G) \tag{7}$$

where $u_{1:|l|}$ denotes the first $|l|$ words of a complete sentence $u$, and $p(u|G)$ is the Viterbi likelihood of $u$.

### 4.4. Maximum Likelihood Estimation for Prediction

We are interested in finding the best grammar and classifier that give us the most accurate predictions based on the generalized Earley parser. Let $G$ be the grammar, $f$ be the classifier, and $D$ be the set of training examples. The training set consists of pairs of complete or partial data sequence $\boldsymbol{x}$ and the corresponding label sequence $\boldsymbol{y}$ for all the frames in $\boldsymbol{x}$. By merging consecutive labels in $\boldsymbol{y}$ that are the same, we can obtain partial label sentences $l$ and predicted labels $z$. Hence we have $D = \{(\boldsymbol{x}, \boldsymbol{y}, l, z)\}$. The best grammar $G^*$ and the best classifier $f^*$ together minimizes the prediction loss:

$$G^*, f^* = \operatorname*{argmin}_{G,f} \mathcal{L}_{pred}(G, f) \tag{8}$$

where the prediction loss is given by the negative log likelihood of the predictions over the entire training set:

$$\begin{aligned}\mathcal{L}_{pred}(G, f) &= -\sum_{(\boldsymbol{x},z)\in D} \log(p(z|\boldsymbol{x})) \\ &= -\sum_{(\boldsymbol{x},l,z)\in D} (\underbrace{\log(p(z|l, G))}_{\text{grammar}} + \underbrace{\log(p(l|\boldsymbol{x}))}_{\text{classifier}})\end{aligned} \tag{9}$$

Given the intermediate variable $l$, the loss is decomposed into two parts that correspond to the induced grammar and the trained classifier, respectively. Let $u \in \{l\}$ be the complete label sentences in the training set (*i.e.*, the label sentence for a complete sequence $\boldsymbol{x}$). The best grammar maximizes the following probability:

$$\prod_{(z,l)\in D} p(z|l, G) = \prod_{(z,l)\in D} \frac{p(z, l|G)}{p(l|G)} = \prod_{u\in D} p(u|G) \tag{10}$$

where denominators $p(l|G)$ are canceled by the previous numerator $p(z, l^-|G)$, and only the likelihood of the complete sentences remain. Therefore inducing the best grammar that gives us the most accurate future prediction is equivalent to the maximum likelihood estimation (MLE) of the grammar for complete sentences in the dataset. This finding lets us to turn the problem (induce the grammar that gives the best future prediction) into a standard grammar induction problem, which can be solved by existing algorithms, *e.g.*, (Solan et al., 2005) and (Tu et al., 2013).

The best classifier minimizes the second term of Eq. 9:

$$\begin{aligned}f^* &= \operatorname*{argmin}_{f} -\sum_{(\boldsymbol{x},l,z)\in D} \log(p(l|\boldsymbol{x}) \\ &\approx \operatorname*{argmin}_{f} -\sum_{(\boldsymbol{x},\boldsymbol{y})\in D} \sum_{k} y_k \log(\hat{y_k})\end{aligned} \tag{11}$$

where $p(l|\boldsymbol{x})$ can be maximized by the CTC loss (Graves et al., 2006). In practice, it can be substituted by the commonly adopted cross entropy loss for efficiency. Therefore we can directly apply generalized Earley parser to outputs of general detectors/classifiers for parsing and prediction.

# 5. Related Work

**Future activity prediction**. This is a relatively new but important domain in computer vision. (Ziebart et al., 2009; Yuen & Torralba, 2010; Ryoo, 2011; Kitani et al., 2012; Kuderer et al., 2012; Wang et al., 2012; Pei et al., 2013; Walker et al., 2014; Vu et al., 2014; Li & Fu, 2014; Wei et al., 2016; Holtzen et al., 2016; Alahi et al., 2016; Xie et al., 2018; Rhinehart & Kitani, 2017; Ma et al., 2017) predict human trajectories/actions in various settings including complex indoor/outdoor scenes. Koppula, Gupta and Saxena (KGS) (Koppula et al., 2013) proposed a model incorporating object affordances that detects and predicts human activities. Koppula *et al*. (Koppula & Saxena, 2016) later used an anticipatory temporal conditional random field to model the spatial-temporal relations through object affordances. Qi *et al*. (Qi et al., 2017) proposed a spatial-temporal And-Or graph (ST-AOG) for activity prediction.

**Hierarchical/grammar models**. Pei *et al*. (Pei et al., 2013) unsupervisedly learned a temporal grammar for video parsing. Holtzen *et al*. (Holtzen et al., 2016) addressed human intent inference by employing a hierarchical task model. Gupta *et al*. (Gupta et al., 2009) learned a visually grounded storyline from videos. The ST-AOG (Qi et al., 2017) is also a type of grammar model. Grammar-based methods show effectiveness on tasks that have compositional structures.

However, previous grammar-based algorithms take symbolic inputs like the traditional language parsers. This seriously limits the applicability of these algorithms. Additionally, the parser does not provide guidance for either training the classifiers or segmenting the sequences. They also lack a good approach to handle grammatically incorrect label sentences. For example, (Qi et al., 2017) finds in the training corpus the closest sentence to the recognized sentence and applies the language parser afterward. In our case, the proposed parsing algorithm takes sequence data of raw signals as input and generates the label sentence as well as the parse tree. All parsed label sentences are grammatically correct, and a learning objective is formulated for the classifier.

# 6. Human Activity Detection and Prediction

We evaluate our method on the task of human activity detection and prediction. We present and discuss our experiment results on two datasets, CAD-120 (Koppula et al., 2013) and Watch-n-Patch (Wu et al., 2015), for comparisons with state-of-the-art methods and evaluation of the robustness of our approach. CAD-120 is the dataset that most existing prediction algorithms are evaluated on. It contains videos of daily activities that are long sequences of sub-activities. Watch-n-Patch is a daily activity dataset that features forgotten actions. Results show that our method significantly outperforms the other methods for activity prediction.

## 6.1. Grammar Induction

In both experiments, we used a modified version of the ADIOS (automatic distillation of structure) (Solan et al., 2005) grammar induction algorithm to learn the event grammar. The algorithm learns the production rules by generating significant patterns and equivalent classes. The significant patterns are selected according to a context-sensitive criterion defined regarding local flow quantities in the graph: two probabilities are defined over a search path. One is the right-moving ratio of fan-through (through-going flux of path) to fan-in (incoming flux of paths). The other one, similarly, is the left-going ratio of fan-through to fan-in. The criterion is described in detail in (Solan et al., 2005).

The algorithm starts by loading the corpus of activity onto a graph whose vertices are sub-activities, augmented by two special symbols, begin and end. Each event sample is represented by a separate path over the graph. Then it generates candidate patterns by traversing a different search path. At each iteration, it tests the statistical significance of each subpath to find significant patterns. The algorithm then finds the equivalent classes that are interchangeable. At the end of the iteration, the significant pattern is added to the graph as a new node, replacing the subpaths it subsumes. We favor shorter patterns in our implementation.

## 6.2. Experiment on CAD-120 Dataset

**Dataset** The CAD-120 dataset is a standard dataset for human activity prediction. It contains 120 RGB-D videos of four different subjects performing 10 high-level activities, where each high-level activity was performed three times with different objects. It includes a total of 61,585 total video frames. Each video is a sequence of sub-activities involving 10 different sub-activity labels. The videos vary from subject to subject regarding the lengths and orders of the sub-activities as well as the way they executed the task.

**Evaluation metrics** We use the following metrics to evaluate and compare the algorithms. 1) Frame-wise detection accuracy of sub-activity labels for all frames. 2) Future 3s online prediction accuracy. We compute the frame-wise accuracy of prediction of the sub-activity labels of the future 3s (using the frame rate of 14Hz as reported in (Koppula et al., 2013)). The predictions are made online at each frame $t$, *i.e.*, the algorithms only sees frame 0 to $t$ and predicts the labels of frame $t+1$ to $t+\delta t$. 3) Future segment online prediction accuracy. At each frame $t$, the algorithm predicts the sub-activity label of the next video segment.

We consider the overall micro accuracy (P/R), macro precision, macro recall and macro F1 score for all evaluation metrics. Micro accuracy is the percentage of correctly classified labels. Macro precision and recall are the average of precision and recall respectively for all classes.

**Comparative methods** We compare our method with four state-of-the-art methods for activity prediction:

1. KGS (Koppula et al., 2013): a Markov random field model where the nodes represent objects and sub-activities, and the edges represent the spatial-temporal relationships. Future frames are predicted based on the transition probabilities given the inferred label of the last frame.
2. Anticipatory temporal CRF (ATCRF) (Koppula & Saxena, 2016): an anticipatory temporal conditional random field that models the spatial-temporal relations through object affordances. Future frames are predicted by sampling a spatial-temporal graph.
3. ST-AOG + Earley (Qi et al., 2017): a spatial-temporal And-Or graph (ST-AOG) that uses a symbolic context-free grammar to model activities. This sets up a comparison between our proposed method and methods that use traditional probabilistic parsers. Since traditional parsers operate on symbolic data, extra efforts need to be done first to extract symbols from sequence data. In this comparative method, the videos are first segmented and labeled by classifiers; the predictions are then made by the original Earley parser.
4. Bidirectional LSTM (Bi-LSTM): a two-layer Bi-LSTM with a hidden size of 256. For the detection task, the output for each frame input is the sub-activity label. For the future 3s prediction, the LSTM is trained to output the label for frame $t+3$s for an input frame at time $t$. For future segment prediction, it outputs the label of the next segment for an input frame. All three tasks use the same training schemes.
5. Bi-LSTM + generalized Earley parser (our method): the proposed generalized Earley parser applied to the classifier output of the above detection Bi-LSTM. The predictions for the next segments are made according to Section 4.3. The lengths of unobserved segments are sampled from a log-normal distribution for the future 3s prediction.

**Feature extraction** All methods in the experiment use the same publicly available features from KGS (Koppula et al., 2013). These features include the human skeleton features and human-object interaction features for each frame. The human skeleton features are location and distance features (relative to the subjects head location) for all upper-skeleton joints of a subject. The human-object features are spatial-temporal, containing the distances between object centroids and skeleton joint locations as well as the temporal changes.

**Experiment results** We follow the convention in KGS (Koppula et al., 2013) to train on three subjects and test on a new subject with a 4-fold validation. The results for the three evaluation metrics are summarized in Table 1, Table 2 and Table 3, respectively. Our method outperforms the comparative methods on all three tasks. Specifically, the generalized Earley parser on top of a Bi-LSTM performs better than ST-AOG, while ST-AOG outperforms the Bi-LSTM. More discussions are highlighted in Section 6.4.

*Table 1.* Detection results on CAD-120.

| Method | Micro | Macro | | |
|---|---|---|---|---|
| | P/R | Prec. | Recall | F1-score |
| KGS (Koppula et al., 2013) | 68.2 | 71.1 | 62.2 | 66.4 |
| ATCRF (Koppula & Saxena, 2016) | 70.3 | 74.8 | 66.2 | 70.2 |
| Bi-LSTM | 76.2 | 78.5 | 74.5 | 74.9 |
| ST-AOG + Earley (Qi et al., 2017) | 76.5 | 77.0 | 75.2 | 76.1 |
| Bi-LSTM + Generalized Earley | **79.4** | **87.4** | **77.0** | **79.7** |

*Table 2.* Future 3s prediction results on CAD-120.

| Method | Micro | Macro | | |
|---|---|---|---|---|
| | P/R | Prec. | Recall | F1-score |
| KGS (Koppula et al., 2013) | 28.6 | – | – | 11.1 |
| ATCRF (Koppula & Saxena, 2016) | 49.6 | – | – | 40.6 |
| Bi-LSTM | 54.2 | 61.6 | 39.9 | 34.1 |
| ST-AOG + Earley (Qi et al., 2017) | 55.2 | 56.5 | 56.6 | 56.6 |
| Bi-LSTM + Generalized Earley | **61.5** | **63.7** | **58.7** | **59.9** |

*Table 3.* Segment prediction results on CAD-120.

| Method | Micro | Macro | | |
|---|---|---|---|---|
| | P/R | Prec. | Recall | F1-score |
| Bi-LSTM | 31.4 | 10.0 | 12.7 | 10.1 |
| ST-AOG + Earley (Qi et al., 2017) | 54.3 | 61.4 | 39.2 | 45.4 |
| Bi-LSTM + Generalized Earley | **72.2** | **70.3** | **70.5** | **67.6** |

### 6.3. Experiment on Watch-n-Patch Dataset

**Dataset** Watch-n-Patch is an RGB-D dataset that features forgotten actions. For example, a subject might fetch milk from a fridge, pour milk, and leave. The typical action "putting the milk back into the fridge" is forgotten. The dataset contains 458 videos with a total length of about 230 minutes, in which people forgot actions in 222 videos. Each video in the dataset contains 2-7 actions interacted with different objects. 7 subjects are asked to perform daily activities in 8 offices and 5 kitchens with complex backgrounds. It consists of 21 types of fully annotated actions (10 in the office, 11 in the kitchen) interacted with 23 types of objects.

**Feature extraction** We extract the same features as described in (Wu et al., 2015) for all methods. Similar to the previous experiment, the features are composed of skeleton features and human-object interaction features extracted from RGB-D images. The skeleton features include angles between connected parts, the change of joint positions and angles from previous frames. Each frame is segmented into super-pixels, and foreground masks are detected. We extract features from the image segments with more than 50% in the foreground mask and within a distance to the human hand joints in both 3D points and 2D pixels. Six kernel descriptors (Wu et al., 2014) are extracted from these image segments: gradient, color, local binary pattern, depth gradient, spin, surface normals, and KPCA/self-similarity.

**Experiment results** We use the same evaluation metrics as the previous experiment and compare our method to ST-AOG (Qi et al., 2017) and Bi-LSTM. We use the train/test split in (Wu et al., 2015). The results for the three evaluation metrics are summarized in Table 4, Table 5 and Table 6, respectively. Our method slightly improves the detection re-sults over the Bi-LSTM outputs, and outperforms the other methods on both prediction tasks. In general, the algorithms make better predictions on CAD-120, since Watch-n-Patch features forgotten actions and the behaviors are more unpredictable. More details are discussed in Section 6.4.

*Table 4.* Detection results on Watch-n-Patch.

| Method | Micro | Macro | | |
|---|---|---|---|---|
| | P/R | Prec. | Recall | F1-score |
| ST-AOG + Earley (Qi et al., 2017) | 79.3 | 71.5 | 73.5 | 71.9 |
| Bi-LSTM | 84.0 | 79.7 | 82.2 | 80.3 |
| Bi-LSTM + Generalized Earley | **84.8** | **80.7** | **83.4** | **81.5** |

*Table 5.* Future 3s prediction results on Watch-n-Patch.

| Method | Micro | Macro | | |
|---|---|---|---|---|
| | P/R | Prec. | Recall | F1-score |
| Bi-LSTM | 42.1 | **66.6** | **62.6** | **61.8** |
| ST-AOG + Earley (Qi et al., 2017) | 48.9 | 43.1 | 39.3 | 39.3 |
| Bi-LSTM + Generalized Earley | **49.0** | 57.0 | 56.5 | 55.3 |

*Table 6.* Segment prediction results on Watch-n-Patch.

| Method | Micro | Macro | | |
|---|---|---|---|---|
| | P/R | Prec. | Recall | F1-score |
| Bi-LSTM | 21.7 | 11.8 | 23.3 | 14.0 |
| ST-AOG + Earley (Qi et al., 2017) | 29.4 | 28.5 | 18.9 | 19.9 |
| Bi-LSTM + Generalized Earley | **35.6** | **59.2** | **59.3** | **53.5** |

### 6.4. Discussion

**How different are the classifier outputs and the final outputs for detection?** Figure 4 shows some qualitative examples of the ground truth segmentations and results given by different methods. The segmentation results show that the refined outputs are similar with the classifier outputs since the confidence given by the classifiers are often very high.

**How does the generalized Earley parser refine the classifier detection outputs?** When the classifier outputs violate the grammar, two types of refinements occur: i) correction and deletion of wrong labels as shown in 4a; ii) insertion of new labels as shown in 4b. The inserted segments are usually very short to accommodate both the grammar and the classifier outputs. Most boundaries of the refined results are well aligned with the classifier outputs.

**Why do we use two metrics for future prediction?** The future 3s prediction is a standard evaluation criterion set up by KGS and ATCRF. However, this criterion does not tell how well the algorithm predicts the next segment label. i) At any time frame, part of the future 3s involves the current sub-activity for most of the times. ii) If the predicted length of the current sub-activity is inaccurate, the frame-wise inaccuracy drops proportionally, even when the future segment label prediction is accurate. Therefore we also compare against the future segment label prediction because it is invariant to variations in activity lengths.

**How well does the generalized Earley parser perform for activity detection and prediction?** From the results we can see that it slightly improves over the classifier outputs for detection, but significantly outperforms the classifier for predictions. The modifications on classifier outputs (corrections and insertions in Figure 4) are minor but important to make the sentences grammatically correct, thus high-quality predictions can be made.

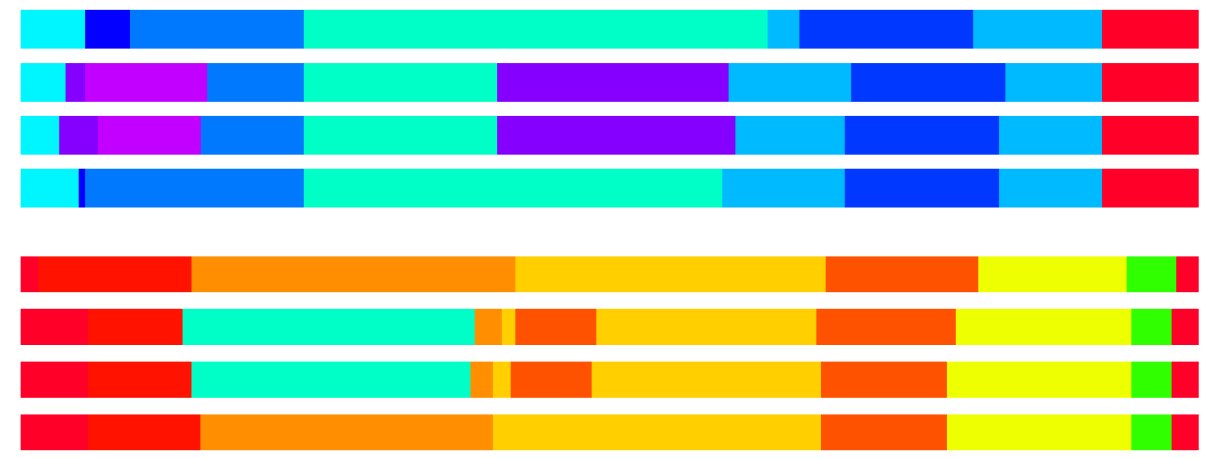

(a) Correction

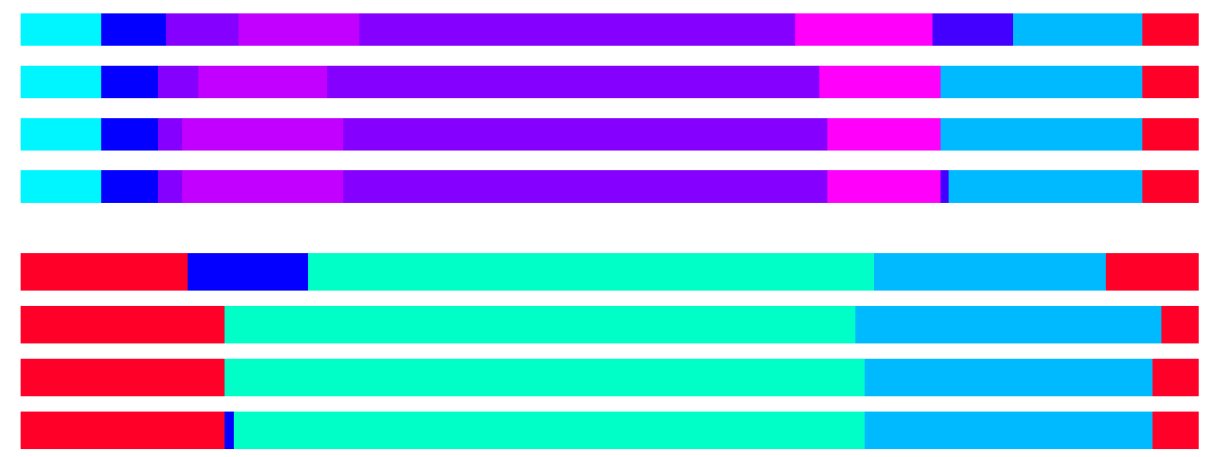

(b) Insertion

*Figure 4.* Qualitative results of segmentation results. In each group of four segmentations, the rows from the top to the bottom show the results of: 1) ground-truth, 2) ST-AOG + Earley, 3) Bi-LSTM, and 4) Bi-LSTM + generalized Earley parser. The results show (a) corrections and (b) insertions by our algorithm on the initial segment-wise labels given by the classifier (Bi-LSTM).

**How useful is the grammar for activity modeling?** From Table 2, Table 3, Table 5 and Table 6 we can see that both ST-AOG and generalized Earley parser outperforms Bi-LSTM for prediction. Prediction algorithms need to give different outputs for similar inputs based on the observation history. Hence the non-Markovian property of grammars is useful for activity modeling, especially for future prediction.

**How robust is the generalized Earley parser?** Comparing Table 3 and Table 6 we can see that there is a performance drop when the action sequences are more unpredictable (in the Watch-n-Patch dataset). But it is capable of improving over the noisy classifier inputs and significantly outperforms the other methods. It is also robust in the sense that it can always find the best sentence in a given language that best explains the classifier outputs.

## 7. Conclusions

We proposed a generalized Earley parser for parsing sequence data according to symbolic grammars. Detections and predictions are made by the parser given the probabilistic outputs from any classifier. We are optimistic about and interested in further applications of the generalized Earley parser. In general, we believe this is a step towards the goal of integrating the connectionist and symbolic approaches.

## Acknowledgment

The authors thank Professor Ying Nian Wu from UCLA Statistics Department for helpful comments on this work. The work reported herein is supported by DARPA XAI N66001-17-2-4029, ONR MURI N00014-16-1-2007, and N66001-17-2-3602.

# Supplementary Material for Generalized Earley Parser

Siyuan Qi [1] Baoxiong Jia [1 2] Song-Chun Zhu [1]

## 1. Review of Formal Grammar

Figure 1 shows an example of English grammar. We review some definitions in the formal language theory:

- Grammar: a set of rules by which valid sentences in a language are constructed.
- Non-terminal: a grammar symbol that can be replaced/expanded to a sequence of symbols.
- Terminal: an actual word in a language; these are the symbols in a grammar that cannot be replaced by anything else. "terminal" is supposed to conjure up the idea that it is a dead-end: pno further expansion is possible.
- Production: a grammar rule that describes how to replace/exchange symbols.

Parsing is the process of analyzing a string of symbols, conforming to the rules of a formal grammar. Figure 2 shows the function of a parsing algorithm: it analyzes a sentence and generates a parse tree.

| | | |
|---|---|---|
| <sentence> | → | <subject> <verbphrase> <object> |
| <subject> | → | "This" \| "Computers" \| "I" |
| <verbphrase> | → | <adverb> <verb> \| <verb> |
| <adverb> | → | "never" |
| <verb> | → | "is" \| "run" \| "am" \| "tell" |
| <object> | → | "the" noun \| "a" noun \| noun |
| noun | → | "university" \| "world" \| "lies" |

*Figure 1.* Production rules of an example English grammar. "|" means "or". Non-terminal symbols are denoted by $< \cdot >$, and terminal symbols are denoted by " · ". Example sentences from the language defined by this grammar are: "This is a university", "Computers run the world", "I never tell lies".

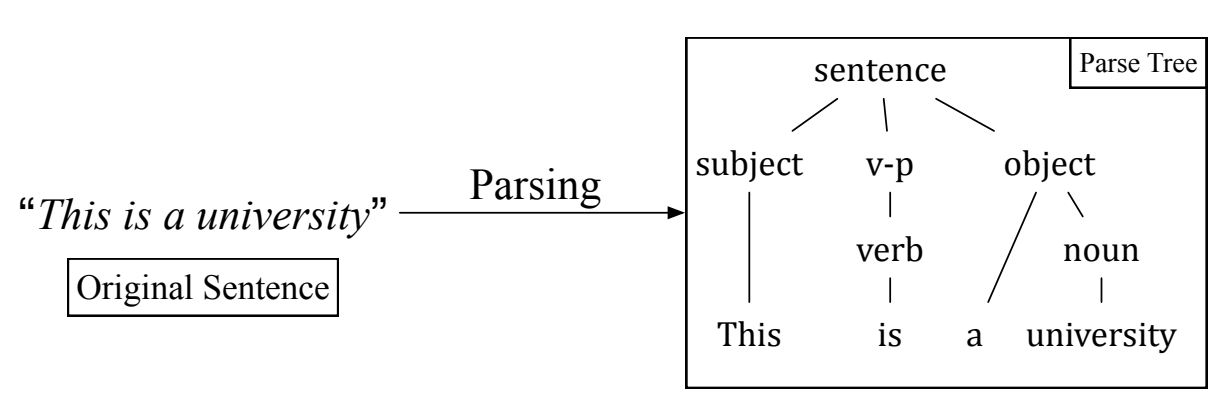

*Figure 2.* Parsing a sentence according to grammar productions.

[1]University of California, Los Angeles, USA [2]Peking University, Beijing, China. Correspondence to: Siyuan Qi <syqi@cs.ucla.edu>.

 

## 2. Example of Generalized Earley Parser

Figure 3 shows a walk-through example of running the generalized Earley parser, which corresponds to Figure 3 in the original paper.. The given language contains three terminal symbols: "0", "1", and "+". The input is a $5 \times 3$ matrix, containing the classification confidence of 3 symbols for 5 frames. The cached probabilities for different prefixes are computed along the inference process. The algorithm essentially expands a prefix tree by three operations: scanning, prediction, and completion. The final parsing output is the optimal string "$0 + 1$" with probability 0.43.

Sample grammar: $\gamma \to R$; $R \to R + R$; $R \to 0|1$

Input (classifier output)

| Frame | 0 | 1 | + |
|---|---|---|---|
| 0 | 0.8 | 0.1 | 0.1 |
| 1 | 0.8 | 0.1 | 0.1 |
| 2 | 0.1 | 0.1 | 0.8 |
| 3 | 0.1 | 0.8 | 0.1 |
| 4 | 0.1 | 0.8 | 0.1 |

Cached probability

| Frame | $\epsilon$ | 0 | 1 | 0 + | 0 + 0 | 0 + 1 | 0 + 1 + |
|---|---|---|---|---|---|---|---|
| 0 | 0 | 8e-1 | 0.1 | 0 | 0 | 0 | 0 |
| 1 | 0 | 6.4e-1 | 1e-2 | 8e-2 | 0 | 0 | 0 |
| 2 | 0 | 6.4e-2 | 1e-3 | 0.58 | 8e-3 | 8e-3 | 0 |
| 3 | 0 | 6.4e-3 | 8e-4 | 6.4e-2 | 5.8e-2 | 0.47 | 8e-4 |
| 4 | 0 | 6.4e-4 | 6.4e-5 | 7e-4 | 1.2e-2 | 0.42 | 4.7e-2 |
| prefix | 1 | 8e-1 | 0.1 | 0.60 | 7.2e-2 | 0.52 | 4.8e-3 |

State: | state # | rule | origin | prefx | comment |

$S(0,0) : l = \epsilon, p(l) = 0.0, p(l...) = 1.0$

| | | | | |
|---|---|---|---|---|
| (1) | $\gamma \to \cdot R$ | $0, 0$ | $\epsilon$ | start rule |
| (2) | $R \to \cdot R + R$ | $0, 0$ | $\epsilon$ | predict: $(1)$ |
| (3) | $R \to \cdot 0$ | $0, 0$ | $\epsilon$ | predict: $(1)$ |
| (4) | $R \to \cdot 1$ | $0, 0$ | $\epsilon$ | predict: $(1)$ |

$S(1,0) : l = \text{"0"}, p(l) = 6.4e-4, p(l...) = 0.8$

| | | | | |
|---|---|---|---|---|
| (1) | $R \to 0 \cdot$ | $0, 0$ | "0" | scan: $S(0,0)(3)$ |
| (2) | $R \to R \cdot + R$ | $0, 0$ | "0" | complete: $(1)$ and $S(0,0)(2)$ |
| (3) | $\gamma \to R \cdot$ | $0, 0$ | "0" | complete: $(2)$ and $S(0,0)(1)$ |

$S(1,1) : l = \text{"1"}, p(l) = 6.4e-4, p(l...) = 0.1$

| | | | | |
|---|---|---|---|---|
| (1) | $R \to 0 \cdot$ | $0, 0$ | "1" | scan: $S(0,0)(4)$ |

$S(2,0) : l = \text{"0 + "}, p(l) = 7.0e-3, p(l...) = 0.599$

| | | | | |
|---|---|---|---|---|
| (1) | $R \to R + \cdot R$ | $0, 0$ | "$0 +$ " | scan: $S(1,0)(2)$ |
| (2) | $R \to \cdot R + R$ | $2, 0$ | "$0 +$ " | predict: $(1)$ |
| (3) | $R \to \cdot 0$ | $2, 0$ | "$0 +$ " | predict: $(1)$ |
| (4) | $R \to \cdot 1$ | $2, 0$ | "$0 +$ " | predict: $(1)$ |

$S(3,0) : l = \text{"0 + 0"}, p(l) = 1.2e-2, p(l...) = 7.2e-2$

| | | | | |
|---|---|---|---|---|
| (1) | $R \to 0 \cdot$ | $2, 0$ | "$0 + 0$" | scan: $S(2,0)(3)$ |

$S(3,1) : l = \text{"0 + 1"}$, **p(l)=0.43**, $p(l...) = 0.52$

| | | | | |
|---|---|---|---|---|
| (1) | $R \to 1 \cdot$ | $2, 0$ | "$0 + 1$" | scan: $S(2,0)(4)$ |
| (2) | $R \to R + R \cdot$ | $0, 0$ | "$0 + 1$" | complete: $(1)$ and $S(2,0)(1)$ |
| (3) | $R \to R \cdot + R$ | $2, 0$ | "$0 + 1$" | complete: $(1)$ and $S(2,0)(2)$ |
| (4) | $\gamma \to R \cdot$ | $0, 0$ | "$0 + 1$" | complete: $(2)$ and $S(0,0)(1)$ |

$S(4,0) : l = \text{"0 + 1 + "}, p(l) = 4.7e-2, p(l...) = 4.8e-2$

| | | | | |
|---|---|---|---|---|
| (1) | $R \to 0 \cdot$ | $2, 0$ | "$0 + 0$" | scan: $S(3,1)(3)$ |

Final output: $l^* = \text{"0 + 1"}$ with probability 0.43

*Figure 3.* An example of the generalized Earley parser. This example corresponds to Figure 3 in the original paper.

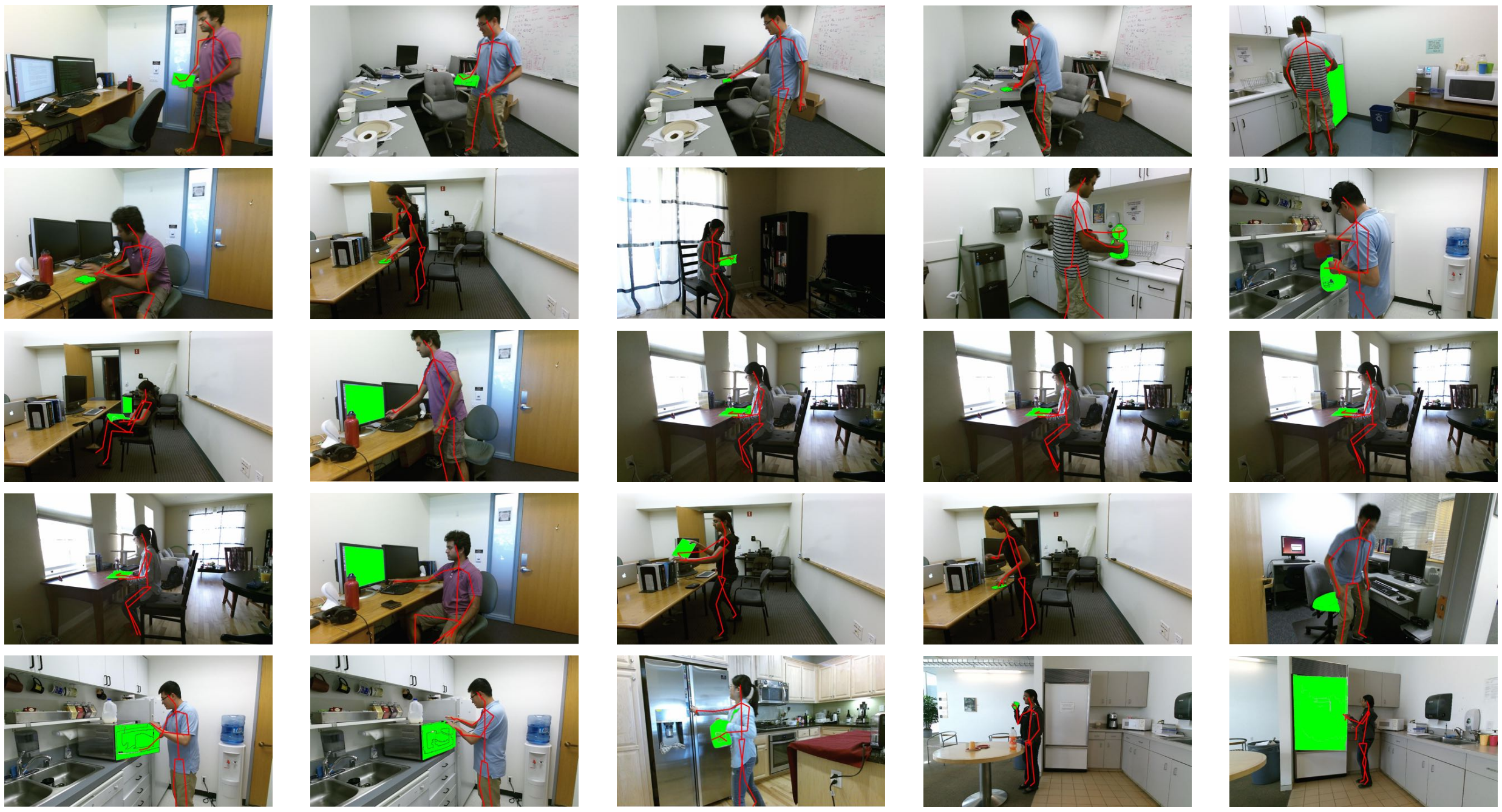

*Figure 4.* Visualization of the selected super-pixels for extracting kernel descriptors

## 3. Feature Extraction

For the CAD-120 dataset, we use the publicly available features from KGS (Koppula et al., 2013). For the Watch-n-Patch dataset, we extract the same features as described in (Wu et al., 2015) for all methods. The features are composed of skeleton features and human-object interaction features. The skeleton features include angles between the connected parts, the change of joint positions and angles regarding previous frames and the first frame. We used zero values for those frames in which the joints are not tracked. The interaction object features are extracted based on RGB-D images. First of all, we calculate the super-pixel segmentation for both RGB and depth images. For each frame, an edge detection algorithm (Dollár & Zitnick, 2013) is applied to both the resized RGB (1920×1080 to 960×540) and the depth images. The selected super-pixels for are shown in Figure 2. Next, we extract super-pixel segmentations from the contour maps generated by the edge detection algorithm. In this case, 0.05 and 0.13 are chosen as the segmenting threshold values for RGB and depth respectively. Second, a moving object detection algorithm is used for monitoring the foreground mask in each frame. Finally, we extract features from the image segments with more than 50% in the foreground mask and within a distance to the human hand joints in both 3D points and 2D pixels. To model the interactive object features more accurately, we removed the false positive segments which overlap with human joints. Overall, we extracted six kernel descriptors (Wu et al., 2014) from these image segments: gradient, color, local binary pattern, depth gradient, spin, surface normals, and KPCA/self-similarity. We first set grid size to 8 and patch size to 16 for generating feature vectors. Then, we construct kernel descriptors by projecting these feature vectors to the visual words generated by K-means algorithm. In this experiment, we generated visual words separately for office scene and kitchen scene. We used 400 visual words for RGB images and 200 for depth images accordingly.

## 4. Segmentation Results

The segmentation results on the Watch-n-Patch are shown in Figure 5. Here we visualize the video segmentation of the ground truth labels, the results generated by ST-AOG (Qi et al., 2017), the Bi-LSTM algorithm, and our algorithm.

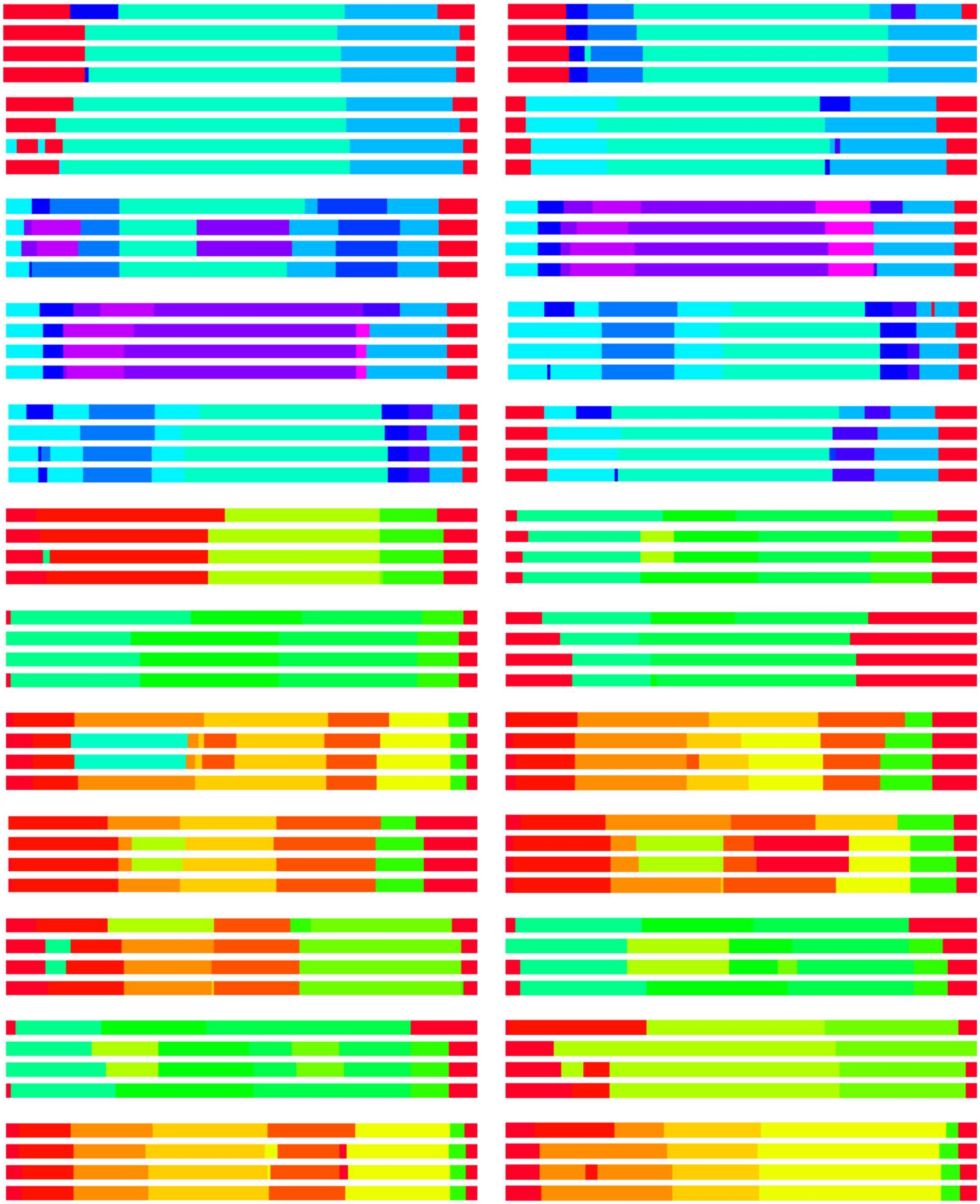

*Figure 5.* Qualitative results of segmentation results. In each group of four segmentations, the rows from the top to the bottom shows: 1) ground-truth, 2) results of ST-AOG, 3) Bi-LSTM, and 4) Bi-LSTM + generalized Earley parser.